\documentclass[10pt,twocolumn,letterpaper]{article}

\usepackage{3dv}
\usepackage{times}
\usepackage{epsfig}
\usepackage{graphicx}
\usepackage{amsmath}
\usepackage{amssymb}
\usepackage{caption}
\usepackage{lipsum}

\usepackage[pagebackref=true,breaklinks=true,colorlinks,bookmarks=false]{hyperref}

\threedvfinalcopy 
\newtheorem{theorem}{Property}
\newtheorem{Lemma}{Lemma}

\usepackage{amsmath}
\DeclareMathOperator*{\argmin}{arg\,min}

\author{David Bensa\"id \\ Technion - Israel Institute of Technology \\ \tt\small dben-said@campus.technion.ac.il  \and Amit Bracha \\ Technion - Israel Institute of Technology \\ \tt\small amit.bracha@cs.technion.ac.il   \and Ron Kimmel \\ Technion - Israel Institute of Technology \\
\tt\small ron@cs.technion.ac.il}
\ifthreedvfinal\fi
\usepackage{pdfpages} 
\begin{document}

\title{Partial Shape Similarity via Alignment of Multi-Metric Hamiltonian Spectra}

\newcommand{\db}[1]{\textcolor{blue}{DB: #1}}
\newcommand{\rk}[1]{\textcolor{red}{RK: #1}}

\newcommand{\az}[1]{\textcolor{blue}{AZ: #1}}
\newcommand{\ab}[1]{\textcolor{cyan}{AB: #1}} 

\twocolumn[{
\maketitle
\begin{center}
    \captionsetup{type=figure}
   \includegraphics[width=1\textwidth]{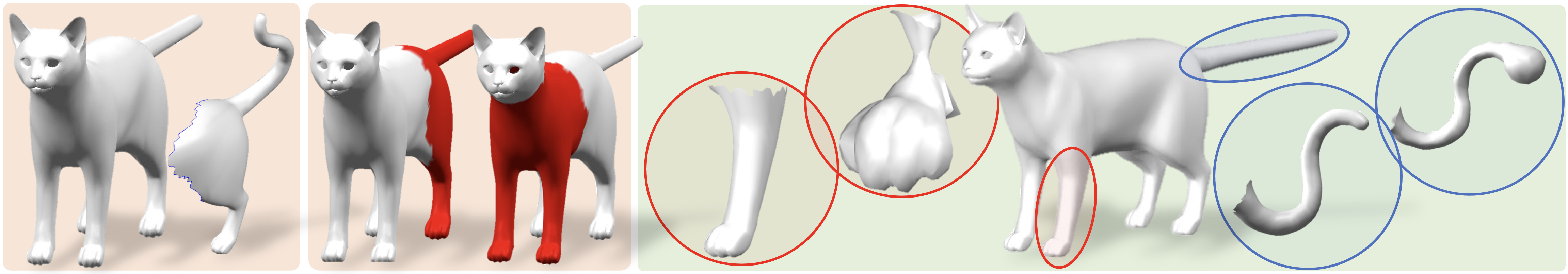}
    \captionof{figure}{
    Illustration depicting the benefit of the proposed multi-metric approach for partial shape matching. 
    \textbf{Light-Red}: Full and part of a non-rigid shape used for the partial shape matching task (left).  
    The partial shape matching results of the proposed algorithm, which considers a single manifold using  
    both the \textit{scale-invariant} (SI) metric and the \textit{regular} metric, and the \cite{rampini2019correspondence} algorithm solution to the same problem using only the regular metric  (Light-Red-Right).
    The proposed method successfully matches the partial shape to the full one. 
    \textbf{Light-Green}: Canonical embedding of the tail and the foreleg (blue and red ellipses, respectively) under the regular metric and the SI metric (left and right circles, respectively). 
    A leg is similar to the tail when considering the regular metric. 
    The SI metric emphasizes the fine geometric discrepancies between a leg and the tail. 
    These fine details explain the correct matching obtained by the multi-metric approach.
    }
\end{center}
}]

\begin{abstract}
Evaluating the similarity of non-rigid shapes with significant partiality is a fundamental task in numerous computer vision applications.
Here, we propose a novel axiomatic method to match similar regions across shapes.
Matching similar regions is formulated as the alignment of the spectra of operators closely related to the Laplace-Beltrami operator (LBO).
The main novelty of the proposed approach is the consideration of differential operators defined on a manifold with multiple metrics. 
The choice of a metric relates to fundamental shape properties while considering the same manifold under different metrics can thus be viewed as analyzing the underlying manifold from different perspectives. 
Specifically, we examine the scale-invariant metric and the corresponding scale-invariant Laplace-Beltrami operator (SI-LBO) along with the regular metric and the regular LBO. 
We demonstrate that the scale-invariant metric emphasizes the locations of important semantic features in articulated shapes. 
A truncated spectrum of the SI-LBO consequently better captures locally curved regions and complements the global information encapsulated in the truncated spectrum of the regular LBO.
We show that matching these dual spectra outperforms competing axiomatic frameworks when tested on standard benchmarks. 
We introduced a new dataset and compare the proposed method with the state-of-the-art learning based approach in a cross-database configuration.
Specifically, we show that, when trained on one data set and tested on another, the proposed axiomatic approach which does not involve training, outperforms the deep learning alternative. 
\end{abstract}

\section{Introduction}
Non-rigid shape matching is a fundamental task encountered in countless 3D computer vision applications such as augmented reality, medical imaging, and facial recognition. 
Matching shapes captured in real-world scenarios, where occlusions and partial views inherently occur, is particularly difficult since the former imperfections usually result in missing geometry.
This challenge has motivated a number of recent papers that deal with \textit{partial shape matching} \cite{rampini2019correspondence, attaiki2021dpfm, litany2017fully, melzi2019zoomout, litany2018partial, rodola2017partial}.

Matching encompasses two related but distinct notions: \textit{correspondence} and \textit{similarity}.
In shape correspondence the matching problem is formulated as a point-wise correspondence between shapes. 
It is generally solved by aligning local descriptors \cite{heider2011local} with a direct \cite{pokrass2013partial} or a functional based \cite{rodola2017partial, litany2017fully, litany2018partial} approach. 
In contrast, shape similarity adopts a global perspective and maps similar \textit{regions} across a pair of shapes. 
In a sense, this approach to the matching problem has more to do with the human visual system, which tends to align shapes by association of similar regions rather than points.
As a result of its global essence, the similarity formulation is resilient to local discrepancies and is thus well suited for challenging settings such as partiality.


Methods evaluating the similarity of shapes usually rely on \textit{shape descriptors}, which characterize a shape or a part of a shape as a whole.
Shape descriptors are generally obtained by aggregating local descriptors \cite{toldo2009bag, pokrass2011correspondence, pokrass2013partial}. 
However, their design requires significant tuning efforts, both on local descriptors and on aggregation methods. 
In 2019, Rampini \textit{et al.} \cite{rampini2019correspondence} addressed partial shape similarity without local descriptors and suggested the spectrum of the Laplace-Beltrami operator (LBO) as a compact shape descriptor.
Similar regions across shapes are then found by formulating and optimizing a spectrum alignment problem.
This approach joins a number of recent efforts that consider the spectrum of the LBO for various tasks \cite{cosmo2019isospectralization, moschella2022learning}.
In this paper, we address partial shape similarity with a framework that does not require solving dense correspondences and essentially does not rely on local descriptors.
The method we introduce generalizes the approach proposed by Rampini \textit{et al.} \cite{rampini2019correspondence}, and considers the spectra of two differential operators defined on a \textit{single} manifold with \textit{multiple} metrics.

The choice of a metric influences fundamental shape properties, including isometry and the notion of similarity itself. 
Considering the same manifold under different metrics can thus be seen as analyzing the underlying shape from different perspectives. 
Specifically, we consider the scale-invariant metric and the spectral decomposition of the scale-invariant LBO \cite{aflalo2013scale} 
along with that of the \textit{regular} LBO.
Roughly speaking, the LBO spectrum captures the support on which it is defined as a whole. 
It has therefore a \textit{global} essence that limits its ability to encapsulate fine details \cite{melzi2018localized, rampini2019correspondence} that are associated with \textit{local} features. 
We prove that the spectrum of the scale-invariant LBO is influenced mainly by curved regions which usually contain meaningful details when considering articulated shapes. 
This property equips the spectrum of the scale-invariant LBO with a particular sensitivity to local curved structures complementing the global shape sensitivity of the spectrum of the regular LBO. 

\noindent
\textbf{The main contributions of this paper include}, 
\begin{itemize}
\item A novel method for partial shape similarity which exploits the complementary perspectives provided by different choices of a metric for the same manifold. 

\item A novel interpretation, with a theoretical support, of the scale-invariant metric as a \textit{prior} on the localization of important details in articulated shapes. 

\item   Significant outperforming the competing axiomatic methods and in particular of the single-spectrum framework tested
on the standard SHREC’16 Partial Matching Benchmark (CUT). 
\item The proposed dual spectra method achieves state-of-the-art results in a cross-dataset setup, compared to a state-of-the-art learning based method. 
\end{itemize}



\section{Related efforts}
Shape matching is a well-established research area and we refer the interested reader to \cite{van2011survey} for a comprehensive survey.
Recently, learning methods \cite{attaiki2021dpfm} demonstrated remarkable results.
Here, we review some recent results related to our line of thought, focusing on \textit{axiomatic} methods for shape matching under partiality.


\paragraph{Spectral methods for partial shape matching.}
Spectral methods, usually based on the spectral decomposition of the Laplace-Beltrami operator, are ubiquitously used in 3D shape analysis \cite{aflalo2015optimality}.
In 2017, Rodol\`a \textit{et al.} proposed the first spectral approach for shape matching under partiality, Partial Functional Map (PFM) \cite{rodola2017partial}. 
PFM extends the seminal framework of functional maps to deal with partiality. 
Partial shape localization and correspondence are alternatively optimized.
Notable follow-up papers include generalizations of PFM to the multipart setting \cite{litany2016non, litany2018partial} and an iterative refinement procedure \cite{melzi2019zoomout} that alternates a coarse matching estimated with PFM and an upsampling strategy.
Finally, Litany \textit{et al.} \cite{litany2017fully} proposed a joint diagonalization method to align the spectral basis of a full and a partial shape in the spectral domain.
Unlike these methods, we were searching for an approach that would not require solving dense correspondences and  would not need local descriptors.

\paragraph{Can you hear the shape of a drum?}
In 1966, Kac published the seminal paper ``Can you hear the shape of a drum?'' \cite{kac1966can}. 
There he formulated a central question in geometry processing: Is it possible to recover the exact geometry of a shape from its spectrum? 

Two decades later, Gordon \textit{et al.}\cite{gordon1992you} answered the last question in the negative with a simple pair of non-isometric 2D polygons presenting identical spectra. 
%
In 2005, Reuters \textit{et al.} \cite{reuter2006laplace} proposed the eigenvalues of the Laplace-Beltrami operator - called \textit{spectrum} - as a global shape signature. 
This signature, named ShapeDNA, is simple, compact, and successfully describes important intrinsic properties of natural shapes such as humans and animals.
In 2007, the introduction of Global Point Signature (GPS) \cite{rustamov2007laplace} marked a shift towards spectral signatures, such as the famous Heat Kernel Signature (HKS) \cite{sun2009concise} and Wave Kernel Signature (WKS) \cite{aubry2011wave}, which also involve the \textit{eigenvectors} of the LBO.
Unlike ShapeDNA, the latter signatures have theoretical guarantees in a continuous perspective and uniquely define shapes up to isometric transformations.
However, local descriptors, like GPS, HKS and WKS, must be associated with aggregation methods, such as the {\it bag-of-words} \cite{toldo2009bag}, to represent an entire shape.
They are therefore less suited than the spectrum for more global tasks such as region localization where larger structures, or \textit{regions}, need to be characterized as a whole.

The eigenvalues of the LBO recently regained a great deal of attention in the geometry processing community.
While known examples of different manifolds with identical spectra exist, they remain specific and concern 2D polygons \cite{gordon1992you} and high-dimensional manifolds \cite{brooks1988constructing, ikeda1980lens,brooks1987isospectral,urakawa1982bounded}.
At the other end, entire classes of manifolds, such as bi-axially symmetric plane domains \cite{zelditch2000spectral}, have been shown to be fully determined by their spectra.
Concerning shapes encountered in real-world scenarios, a line of recent papers based on the spectrum of the LBO demonstrated excellent empirical results.
Cosmo \textit{et al.} \cite{cosmo2019isospectralization} have shown that the spectrum is informative enough to recover shapes in numerous practical cases, and Moschella \textit{et al.} \cite{moschella2022learning} have learned the joint spectrum of a set of partial shapes without computing the 3D geometry of their union.
The basis of our method is the recent approach of Rampini \textit{et al.} \cite{rampini2019correspondence} who formulated the shape localization problem as a spectrum alignment problem.

\paragraph{Self-functional map.} In 2018, Halimi \textit{et al.} \cite{halimi2018self} proposed the {\it self-functional map} framework for shape retrieval. 
It was later extended to shape correspondence  \cite{bracha2020shape}. 
Self-functional maps are compact shape representations that characterize shapes by the inner product of eigenfunctions of Laplace-Beltrami operators induced by two different metrics.
The approach of Halimi \textit{et al.} is therefore closely related to our line of thought since it also relies on the analysis of a \textit{single} shape with \textit{multiple} metrics. 
\section{Background: Laplace-Beltrami and Hamiltonian operators}
\label{sec: Background}
\paragraph{Shapes as Riemannian manifolds.}
We model a shape as a Riemannian manifold $\mathcal{M} = (S,g)$, where $S$ is a smooth two-dimensional manifold embedded in $\mathbb{R}^3$ and $g$ a metric tensor, also referred to as \textit{first fundamental form}.
The metric tensor can be used to define \textit{geometric} quantities on the manifold, such as lengths of curves, and angles between vector fields, and distances between points. 
Consider a parametric surface $S(u,v): \Omega \subseteq \mathbb{R}^2 \rightarrow \mathbb{R}^3$, equipped with the regular intrinsic metric $g$,
\begin{eqnarray}
(g_{ij}) &=& \begin{pmatrix} \langle S_u, S_u \rangle  & \langle S_u, S_v \rangle \\ \langle S_v, S_u \rangle  & \langle S_v, S_v \rangle \end{pmatrix}.
\end{eqnarray}
An infinitesimal length element $ds$ on the surface $S$ can then be defined by,
\begin{eqnarray}
ds^2 &=& \textbf{du}^T (g) \, \textbf{du} \,,
\end{eqnarray}
with $\textbf{du} \triangleq \begin{pmatrix} du \\ dv \end{pmatrix}$.
Formally, $\mathcal{M}$ is a Riemannian manifold that assigns to every point ${m \in \mathcal{M}}$ a tangent plane $T_{m}\mathcal{M}$ and an inner product ${\langle.\rangle_{g} : T_{m}\mathcal{M} \times T_{m}\mathcal{M} \rightarrow \mathbb{R}}$.
\subsection{Laplace-Beltrami operator}
The Laplace-Beltrami operator (LBO) ${\Delta_{g}}$ is an ubiquitous tool in shape analysis that earned the title of “Swiss army knife” in the geometry processing community.
The LBO is a generalization of the regular Laplacian operator to Riemannian manifolds.
Formally, 
\begin{eqnarray}
\Delta_g f \triangleq - \frac{1}{\sqrt{|g|}} \text{div} (\sqrt{|g|} g^{-1} \nabla f) \, , \, \, f \in \mathcal{L}^2(\mathcal{M}) \,,
\label{eq: LBO}
\end{eqnarray}
where ${\mathcal{L}^2(\mathcal{M})}$ stands for the Hilbert space of square-integrable scalar functions defined on $\mathcal{M}$.

\paragraph{Spectral analysis.} 
The semi-positive definite LBO admits the following spectral decomposition with homogeneous Dirichlet boundary conditions,
\begin{align}
\Delta_{g} \phi_i(x) &= \lambda_{i} \phi_i(x) \, ,  & x &\in \mathcal{M} \setminus \partial \mathcal{M}\cr
\phi_i(x) &= 0 \, , & x &\in \partial \mathcal{M}
\end{align}
where $\partial \mathcal{M}$ stands for the boundary of manifold $\mathcal{M}$.
The basis ${\{ \phi_i\}_{i\geq 0}}$ defined by the LBO spectral decomposition is invariant to isometries that may include non-rigid transformations. 
It is commonly interpreted as a generalization of the Fourier basis on flat domains to Riemannian manifolds \cite{taubin1995signal}. Truncating the basis has been proven to be unique and optimal in approximating smooth functions on a given shape, by which the basis is defined \cite{aflalo2015optimality}.. 


\paragraph{Discretization.} 
In a discrete setting, $S$ is
approximated by a triangulated mesh with $n$ vertices. The discrete LBO can then be approximated by,
\begin{eqnarray}
    \textbf{L} &=& \textbf{A}^{-1} \textbf{W} \,,
\end{eqnarray}
where $\textbf{A} \in \mathbb{R}^{n \times n}$ is known as the {\it mass matrix} containing an area element about each vertex, while $\textbf{W} \in \mathbb{R}^{n \times n}$ is the {\it cotangent weight matrix} \cite{pinkall1993computing}.
The spectral decomposition of the discrete LBO can be computed as a solution for the generalized eigenvalue problem,
\begin{align}
\textbf{W} \phi_i(x)  &= \lambda_{i} \textbf{A} \phi_i (x) \, ,  & x &\in \mathcal{M} \setminus \partial \mathcal{M} \cr
\phi_i(x) &= 0 \, , & x &\in \partial \mathcal{M}
\end{align}
with homogeneous Dirichlet boundary conditions.

\subsection{The Hamiltonian operator in shape analysis}
\begin{figure}[htbp]
    \centering
    \includegraphics[width=0.5\textwidth]{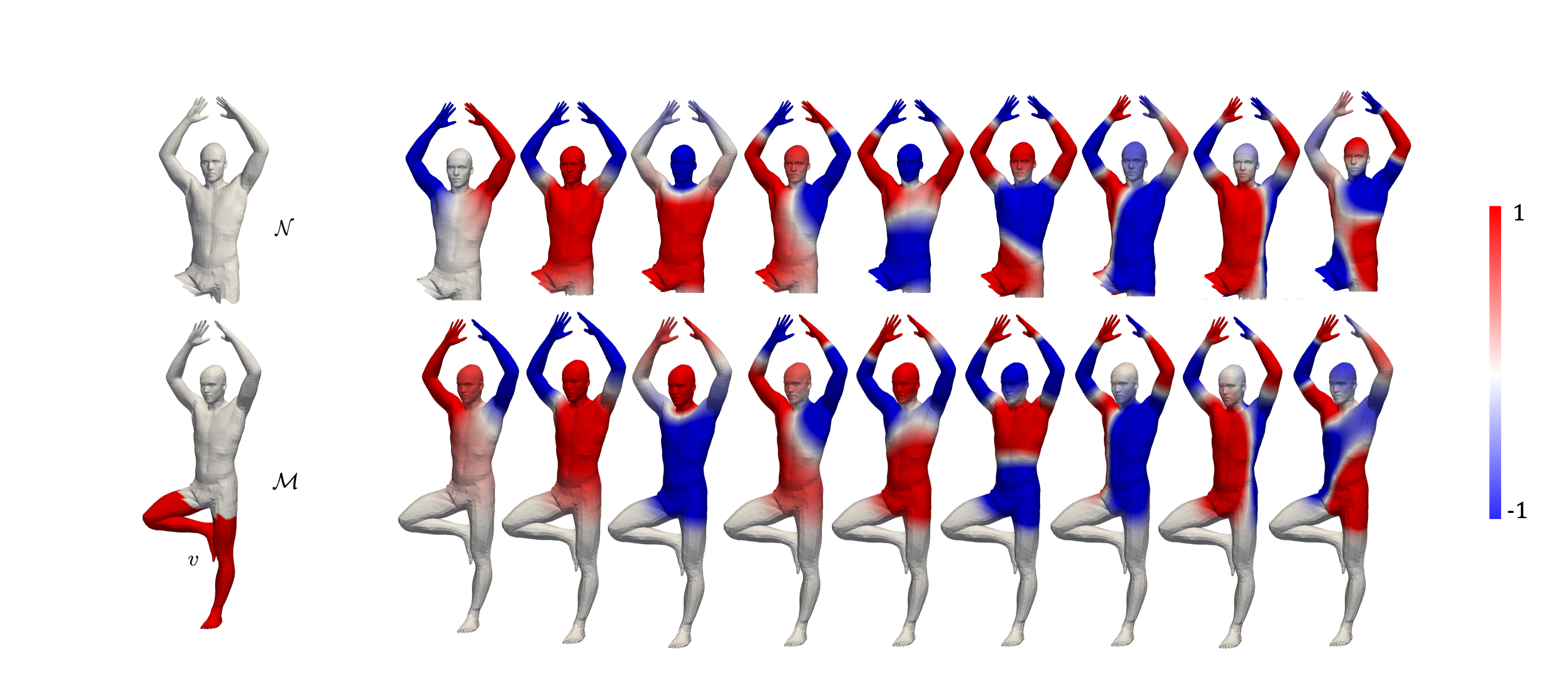}
    \caption{Top: First eigenvectors of the LBO of the \textit{partial} shape $\mathcal{N}$. Bottom: First eigenvectors of the Hamiltonian of the \textit{full} shape $\mathcal{M}$. The Hamiltonian is defined with a step potential $v$ (in red) corresponding to the effective support of $\mathcal{N}$. With the adapted potential, the eigenfunctions of the LBO and the Hamiltonian are similar up to their sign.}
    \label{fig: Shape_Hamiltonian}
\end{figure}
In 2018, Choukroun \textit{et al.} \cite{choukroun2018hamiltonian} adapted the well-known Hamiltonian operator from quantum mechanics to shape analysis.
The Hamiltonian naturally appears in the celebrated Schr\"{o}dinger equation that describes the \emph{wave function} of a particle, i.e., its spatial and temporal probability distribution,
\begin{eqnarray}
     i\hbar \frac{\partial}{\partial t} \Psi(x,y,t) &=& \left (\frac{\hbar}{2m} \Delta + \text{V}(x,y,t)\right )  \Psi(x,y,t) \cr \cr
     &=& \text{H} \Psi(t,x,y),
\label{eq: Schrodinger}
\end{eqnarray}
where $\text{H}$ is the {\it Hamiltonian} operator, $\hbar$  the Plank constant, and $\text{V}$ a scalar function mapping the domain  ${\Omega}$ on which the Schr\"{o}dinger equation is defined.
In the case of a Riemannian manifold, $\mathcal{M}$, and assuming a time-independent step-potential, the Hamiltonian in Eq. (\ref{eq: Schrodinger}) becomes,
\begin{eqnarray}
    \text{H}_{g} \triangleq \Delta_{g} + v,
\end{eqnarray}
with $v: \mathcal{M} \rightarrow \mathbb{R}$.
It is a semi-positive definite operator that admits a spectral decomposition,
\begin{align}
(\Delta_{g} + v) \phi_i (x)  &= \lambda_{i} \phi_i (x) \, ,  & x &\in \mathcal{M} \setminus \partial \mathcal{M}  \cr
\phi_i(x) &= 0 \, , & x &\in \partial \mathcal{M}
\end{align}
with Dirichlet boundary conditions.
Here, ${\lambda_i}$ is known as the \emph{energy} of the \emph{wave function} $\phi_i$. 
Using the notations introduced in Section \ref{sec: Background}, a discrete versions of the Hamiltonian $\textbf{H}$ and of the related eigendecomposition problem are respectively,
\begin{eqnarray}
\textbf{H} &=& \textbf{A}^{-1}\textbf{W} + \textbf{V},
\end{eqnarray}
\begin{align}
(\textbf{W} + \textbf{A} \textbf{V}) \phi_i(x) &= \lambda_i \textbf{A} \phi_i(x) \, , & x &\in \mathcal{M} \setminus \partial \mathcal{M}  \cr
\phi_i(x) &= 0 \, , & x &\in \partial \mathcal{M}, 
\label{eq: discrete hamiltonian}
\end{align}
where $\textbf{V} \in \mathbb{R}^{n \times n}_{+}$ is a diagonal matrix containing the values of the potential $v$ at each vertex of the discretized manifold.
We refer the interested reader to \cite{choukroun2018hamiltonian} for a comprehensive survey of the Hamiltonian operator in the context of shape analysis.
For our discussion, the most important property of the Hamiltonian is,
\begin{theorem}
{Let $\mathcal{M} $ be a Riemmanian manifold and  ${v: \mathcal{M} \rightarrow \mathbb{R}_{+}}$ a potential function. 
The eigenfunction $\phi_i$ of the Hamiltonian exponentially decays in every ${\hat{s} \in S}$ such that ${v(\hat{s}) > \lambda_i}$.}
\label{theorem: hamiltonian}
\end{theorem}
When it comes to shapes, the potential can be considered as a mask determining the domain at which the LBO embedded in the Hamiltonian is effective (see Figure \ref{fig: Shape_Hamiltonian}).
A second key property of the Hamiltonian is the differentiability of its eigenvalues with respect to its potential function \cite{abou20091}. 
\begin{theorem}
{The eigenvalues $\{ \lambda_i \}_{i\geq0}$ of the discretized Hamiltonian operator $\mathbf{H}$ are differentiable with respect to the potential ${v}$.
Namely,
\begin{eqnarray}
\frac{\partial \lambda_i}{\partial v} &=&  
\mathbf{A} (\phi_i \otimes \phi_i) \,,
\end{eqnarray}
where ${\otimes}$ stands for the element-wise multiplication, and ${\phi_i}$ is the eigenvector corresponding to $\lambda_i$.}
\label{theorem: derivation}
\end{theorem}
\section{Scale-invariant geometry}
The search for meaningful representations of images and shapes lays at the heart of computer vision and geometry processing.
These representations should be \textit{invariant} to defined classes of transformation such as Euclidean transformations, namely translations, rotations, and scaling.
In 2013, Aflalo \textit{et al.} \cite{aflalo2013scale} introduced the scale-invariant metric that we summarize next.
We refer the reader to the supplementary for a detailed presentation of the scale-invariant metric.

\subsection{Scale-invariant metric}
\label{subsec: scale invariant geometry}
The scale-invariant metric for surfaces is based on the Gaussian curvature K \cite{do2016differential}, which is intuitively related to the \textit{bending} of the surface.
The Gaussian curvature depends on the scale of the shape and the modulation of the Euclidean arc length by K consequently leads to a \textit{scale-invariant} metric.
Using the notations introduced in Section \ref{sec: Background}, a scale invariant pseudo-metric $\tilde{g}$ can be defined as,
\begin{eqnarray}
    \tilde{g}_{ij} \triangleq |K| \, g_{ij} \,,
\label{eq: scale invariant metric}
\end{eqnarray}
where $g$ is the \textit{regular} first fundamental form. 
Remark that the addition of a small positive constant $\epsilon$ to the Gaussian curvature in \ref{eq: scale invariant metric} prevents the expression from vanishing and allows the definition of a formal metric.
The modulation by the Gaussian curvature shrinks any intrinsically flat region into a point. 
The LBO of a Riemmanian manifold $\tilde{\mathcal{M}}$ equipped with the scale-invariant metric is called the \textit{scale-invariant Laplace-Beltrami Operator} (SI-LBO) \cite{aflalo2013scale}.
A closed form of the SI-LBO is derived by plugging the scale-invariant metric $\tilde{g}$ instead of the \textit{regular} metric $g$ into the definition of the LBO,
\begin{eqnarray}
\Delta_{\tilde{g}} f &=&- \frac{1}{\sqrt{|\tilde{g}|}} \text{div} (\sqrt{|\tilde{g}|} \tilde{g}^{-1} \nabla f) \, , \, \, f \in L^2(\tilde{\mathcal{M}}) \,. \cr & &
\end{eqnarray}
The discrete version of the SI-LBO and of the related eigendecomposition problem are respectively, 
\begin{eqnarray}
\tilde{\textbf{L}} &=& |\textbf{K}|^{-1}\textbf{A}^{-1}\textbf{W} \,,
\end{eqnarray}
\begin{align}
(\textbf{W} + \textbf{A}|\textbf{K}| \textbf{V}) \phi_i(x) &= \lambda_i \textbf{A} |\textbf{K}|   \phi_i(x) \,, & x &\in \mathcal{M} \setminus \partial \mathcal{M} \cr
\phi_i(x) &= 0 \, , & x &\in \partial \mathcal{M} \,, 
\label{eq: discrete SILBO}
\end{align}
where $\textbf{K} \in \mathbb{R}^{n \times n}$ contains the discrete curvature at the vertices of $\mathcal{M}$ along its diagonal. 
\subsection{Scale-invariance as a shape prior}
\label{subsec: Scale-invariance as a shape prior}
\begin{figure}[htbp]
    \includegraphics[width=0.5\textwidth]{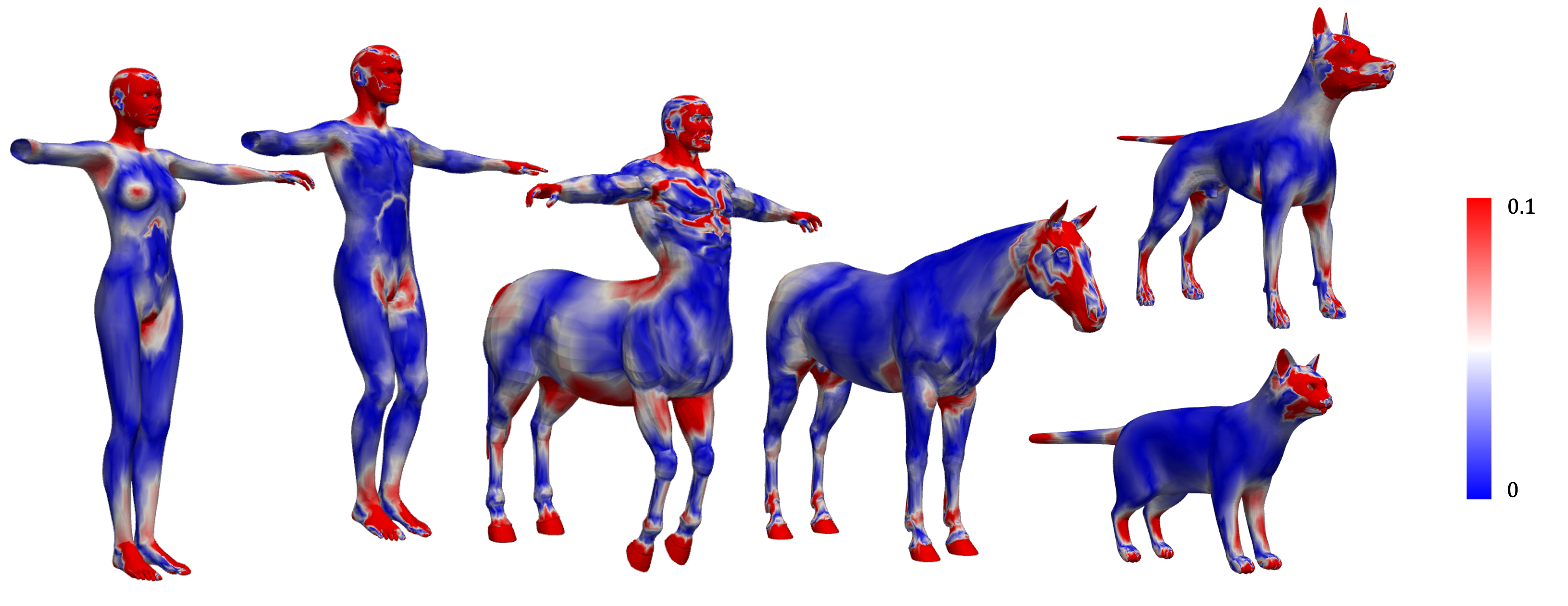}
    \caption{Absolute value of the Gaussian curvature on shapes from TOSCA \cite{bronstein2008numerical}. 
    The curvature is smoothed for better visualization. Regions with large curvature (in red) have a larger influence on the spectrum of the SI-LBO according to Property \ref{theorem: perturbation analysis}.}
    \label{fig: prior}
\end{figure}
Consider the following interpretation of the  scale-invariant metric. 
The scale-invariant metric can be apprehended as a \textit{prior} on the localization of important semantic features in articulated shapes such as humans and most animals. 
Namely, key elements are assumed to be localized in curved regions. 
Figure \ref{fig: prior} shows the curvature on various shapes from SHREC'16 \cite{cosmo2016shrec}.
In the human case for instance, the scale invariant metric \textit{accentuates} the head, the hands and the feet, at the expense of flat regions such as the back and the legs.
Intuitively, the scale-invariant metric shrinks intrinsically flat regions into points. 
Consequently, it focuses on curved regions, see the Supplementary.
We formalize this intuition in the spectral domain with a direct generalization of the Weyl law for the SI-LBO.
\begin{Lemma}
Let $\tilde{\mathcal{M}} = (S: \Omega \subseteq \mathbb{R}^2 \rightarrow \mathbb{R}^3 , \tilde{g})$ be a Riemmanian manifold and $\{\tilde{\lambda}_i \}_{i\geq 1}$ the spectrum of the scale-invariant LBO of $\tilde{\mathcal{M}}$. It holds,
\begin{eqnarray*}
   \tilde{\lambda}_i &\sim& \frac{2\pi i}{\int_{\Omega}  |K| da} \cr
    \textup{or,  }\tilde{\lambda}_i &\sim& \frac{2\pi i}{\textup{trace} (|\textbf{K}| \textbf{A})} \, \textup{ in the discrete setting.}
\end{eqnarray*}
\label{theorem: SI-Weyl}
\end{Lemma}
\noindent \textbf{Proof (discrete setting).}
Remark that, in the spectral domain, the introduction of the scale-invariant metric boils down to the substitution of the area matrix $\textbf{A}$, that weighs the vertices according to their area, by ${|\textbf{K}|\textbf{A}}$. 

The proof in the continuous setting is presented in the Supplementary. A simple perturbation analysis expresses the influence of the curvature on the spectrum of the SI-LBO.
\begin{theorem}
Consider a \textit{small} perturbation $\delta_p$ of the curvature at $p \in \tilde{\mathcal{M}}$.
Denote by $\delta K \triangleq \frac{\delta_p}{\textup{trace}(|\textup{K}|\textup{A})}$ the relative perturbation of the curvature and by $\delta \tilde{\lambda} \triangleq \frac{\tilde{\lambda}_i - \tilde{\mu}_i}{\tilde{\lambda}_i}$  
the relative perturbation of $\tilde{\lambda}_i$, where $\tilde{\mu}_i$ designates the $i^{th}$ eigenvalue of the perturbed manifold. $ \delta \tilde{\lambda}$ respects, 
\begin{eqnarray*}  
    \delta \widetilde{\lambda} \sim \delta K \,.
\end{eqnarray*}
\label{theorem: perturbation analysis}
\end{theorem}
\noindent \textbf{Proof.}
From Lemma \ref{theorem: SI-Weyl},
\begin{eqnarray*}
    \tilde{\lambda}_i \sim \frac{2\pi i}{\text{trace}(|\textbf{K}| \textbf{A})},   \quad \tilde{\mu}_i \sim \frac{2\pi i}{\text{trace}(|\textbf{K}| \textbf{A}) + \textbf{A}_p \delta_p} \,.
\end{eqnarray*}
\begin{eqnarray*}
\text{Therefore, } \delta \tilde{\lambda} &\triangleq& \frac{\tilde{\lambda}_i - \tilde{\mu}_i}{\tilde{\lambda}_i} 
    \cr
    &\sim& \frac{1}{\tilde{\lambda}_i} \frac{2\pi i \textbf{A}_p \delta_p \textbf{A}_p}{\text{trace}(|\textbf{K}| \textbf{A}) (\text{trace}(|\textbf{K}| \textbf{A}) + \textbf{A}_p \delta_p)} 
    \cr
    &\sim&  \frac{1}{\tilde{\lambda}_i} \frac{2\pi i}{\text{trace}(|\textbf{K}| \textbf{A})} \frac{\textbf{A}_p \delta_p}{\text{trace}(|\textbf{K}| \textbf{A})} \cr
    &\sim& \textbf{A}_p \delta \textbf{K}
\end{eqnarray*}

Property \ref{theorem: perturbation analysis} implies that the spectrum of the SI-LBO is mostly determined by curved regions.
Empirically, this property also extends to the modes of the SI-LBO and the basis induced by the spectral decomposition of the SI-LBO is more concentrated, or \textit{compressed}, in curved regions (see Supplementary). 
Modes related to the first eigenvalues of the SI-LBO capture fine structures, such as fingers.
An interesting parallel can thus be drawn between the SI-LBO and differential operators with \textit{compressed modes} \cite{ozolicnvs2013compressed, choukroun2018hamiltonian, melzi2018localized} tailored to describe local functions on shapes.
Unlike those operators, the SI-LBO has the key advantage to be unsupervised and to determine axiomatically the regions where localized spectral analysis should be performed.

\section{Method}
We propose a framework to find the effective support of a partial shape in a full shape.

\noindent \textbf{Overview.} Using Property \ref{theorem: hamiltonian}, we reduce the localization of a \textit{partial} shape within a \textit{full} shape to a search for a Hamiltonian's potential corresponding to the effective support of the partial shape.
The search for the potential function is formulated with a cost function promoting the alignment of the spectra of the Hamiltonian defined on the full shape and the LBO defined on the partial shape.
Finally, Property \ref{theorem: derivation} allows the minimization of the cost function with a first-order optimization algorithm.
Although the problem of aligning spectra is non-convex, Rampini \textit{et al.} \cite{rampini2019correspondence} have empirically shown that known optimization algorithms can be used to find \textit{good} local minima. 

\paragraph{A dual look on a single manifold.} 
We jointly process Riemannian manifolds representing a \textit{single} manifold equipped with \textit{multiple} metrics. 
Essential shape properties, including the notion of similarity itself, are affected by the choice of a metric and a pair of manifolds can be both isometric and non-isometric according to different metrics \cite{halimi2018self}. 
Considering multiple metrics can therefore be viewed as considering alternative perspectives of the same manifold, each being sensitive to distinct types of deformation.
Specifically, we use the \textit{regular} and the scale-invariant metrics.
As discussed earlier (see Section \ref{subsec: Scale-invariance as a shape prior}), the scale-invariant spectrum incorporates local information from key regions that complement the global perspective of the regular spectrum.

\paragraph{Problem formulation.} We formulate the partial shape similarity problem as a region localization task.
We consider a full and a partial manifold denoted respectively by ${S_f}$ and ${S_p}$. 
They define a full and a partial shape up to a metric choice. 
The output of the proposed framework is an indicator function indicating the \textit{region} of the full shape corresponding to the partial shape.
Formally, the output scalar function ${o: S_f \rightarrow \{1, 0\}}$ should respect,
\begin{eqnarray}
    S^{'} \sim  S_p
\end{eqnarray}
where ${S^{'}=\{x \in S_f | \, o(x)=1\}}$ and ${\sim}$ stands for an isometry relation according to the \textit{regular} Riemannian metric.

\paragraph{Notations.} We denote the full shape equipped with the \textit{regular} metric by ${\mathcal{M}=(S_f, g)}$ and the spectrum of the Hamiltonian defined over ${\mathcal{M}}$ by ${\{\lambda_i\}^{k}_{i=1}}$ with ${\lambda_1 \leq ... \leq \lambda_k}$.
${\, \tilde{\mathcal{M}}=(S_f, \tilde{g})}$ stands for the full shape defined with the scale-invariant metric and ${\{\tilde{\lambda}_i\}^{k}_{i=1}}$, with ${\tilde{\lambda}_1 \leq ... \leq \tilde{\lambda}_k}$, for the spectrum of the \textit{scale-invariant Hamiltonian}.
We denote by ${\Phi \in \mathbb{R}^{n \times k}}$ and ${\tilde{\Phi} \in \mathbb{R}^{n \times k}}$ the $k$ first eigenfunctions of the Hamiltonian and of the scale-invariant Hamiltonian of $\mathcal{M}$.
In the same way, the partial shape equipped with the \textit{regular} and the scale-invariant metrics are referred to as ${\mathcal{N}=(S_p, g), \, \tilde{\mathcal{N}}=(S_p, \tilde{g})}$.
The spectra of the LBO and of the SI-LBO are respectively denoted by ${\{\mu_i\}^{k}_{i=1}}$ and by ${\{\tilde{\mu}_i\}^{k}_{i=1}}$, with ${\mu_1 \leq ... \leq \mu_k}$ and ${\tilde{\mu}_1 \leq ... \leq \tilde{\mu}_k}$.

\paragraph{Cost function.}
We consider a cost function promoting the alignment of the spectra of the LBO and of the SI-LBO of $\mathcal{N}$ with the the spectra of the regular and the scale-invariant Hamiltonian of $\mathcal{M}$. Namely,
\begin{eqnarray}
    f(V)&=& \|\lambda(V) -\mu \|^{2}_{w} + \|\tilde{\lambda}(V) - \tilde{\mu}\|^{2}_{w} \,.
\label{eq: cost function}
\end{eqnarray}
Following \cite{rampini2019correspondence}, the weighted L2 norm ${\|.\|_w}$ is defined as,
\begin{eqnarray}
    \|a - b\|^{2}_{w} &=& \sum^{k}_{i=1} \frac{1}{b_{i}^{2}} (a_i - b_i)^2 \,,
\end{eqnarray}
to mitigate the weight given to high frequencies.

\paragraph{Optimization.} The cost function Eq. (\ref{eq: cost function}) induces a constrained optimization problem, 
\begin{eqnarray}
    \argmin_{\text{V} \geq 0} f(V) \,.
\label{eq: constrained optimization}
\end{eqnarray}
According to Property \ref{theorem: derivation}, the gradient of the last equation with respect to $v$ is,
\begin{eqnarray}
    \nabla_{v}f &=& 2 \, (\Phi \otimes \Phi) ((\lambda - \mu) \oslash \mu^2) \cr 
    &&\,+ \,2 \, (\tilde{\Phi} \otimes \tilde{\Phi}) ((\tilde{\lambda} - \tilde{\mu}) \oslash \tilde{\mu}^2) \,.
\end{eqnarray}
To simplify the optimization process, we minimize an unconstrained relaxation of Eq. (\ref{eq: constrained optimization}) instead,
\begin{eqnarray}
\argmin_{V} f(q(V)) \,,
\label{eq: optimization}
\end{eqnarray}
with $q: \mathbb{R} \rightarrow \mathbb{R}_{+}$ a smooth function acting element-wisely.
We consider ${q_1(x) = x^2}$ as well as $q_2(x) = \text{c} (\text{tanh}(v) + 1)$ with $\text{c} \gg \mu_k$.
By promoting high step potentials, the saturation function $q_2$ \cite{rampini2019correspondence} limits the eigenfunctions that can be considered within the region where $v \approx 0$.
Eq. (\ref{eq: optimization}) is finally minimized with a trust-region method, a classical first order optimization algorithm.

\section{Experiments}
\begin{figure}[htbp]
    \centering
    \includegraphics[width=0.5\textwidth]{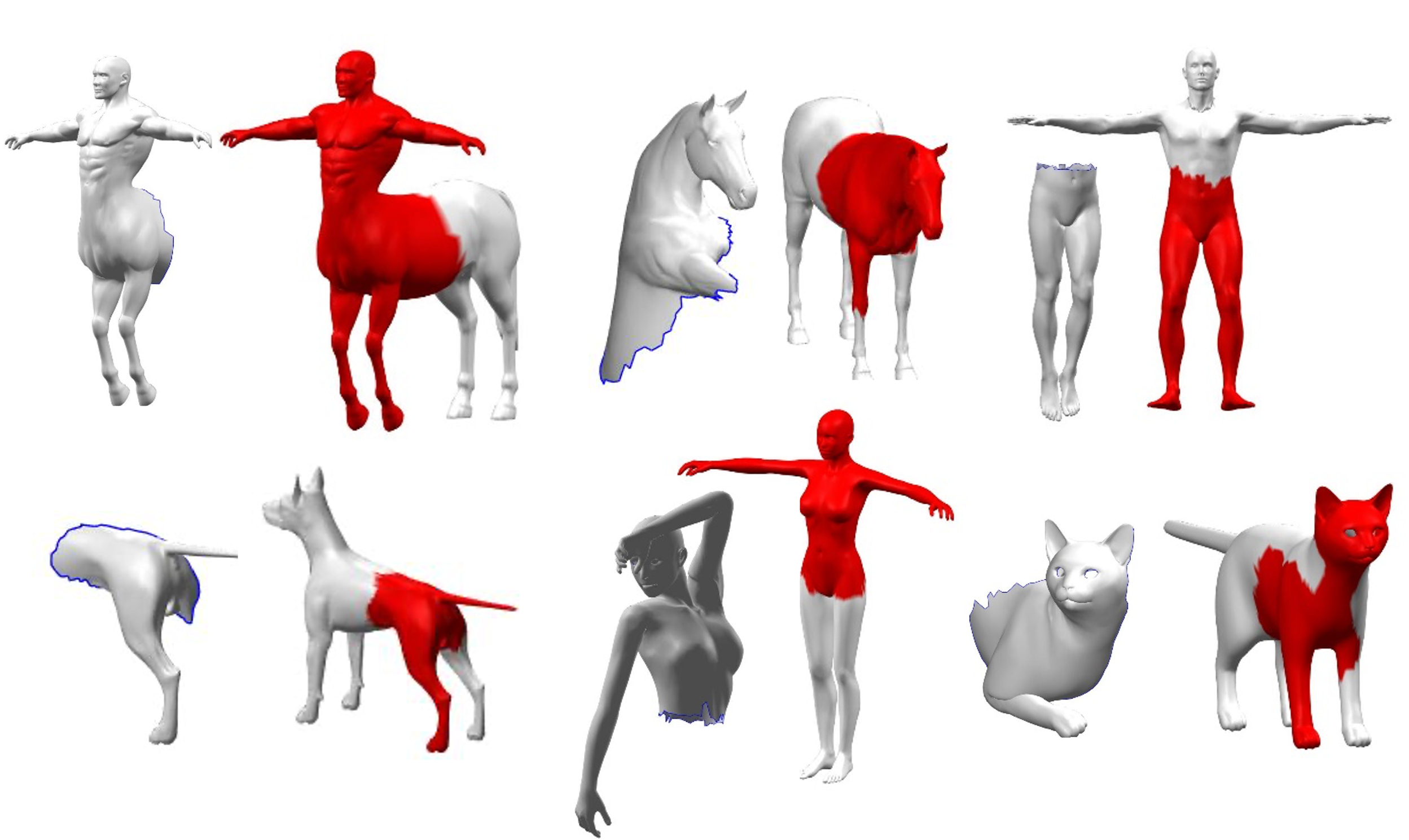} 
    \caption{Successful examples of the proposed method on shapes from SHREC'16 \cite{cosmo2016shrec}. Red regions on the full shapes indicate the regions associated with queried parts.} 
    \label{fig: qualitative results}
\end{figure}

\begin{table}[htbp]
\begin{center}
\begin{tabular}{|c|c|} 
\hline
Method & \multicolumn{1}{c|}{mean IoU} \\
\hline
\hline
Bag of words based on SHOT \cite{toldo2009bag} & \multicolumn{1}{c|}{$0.453$}  \\
\hline
PFM \cite{rodola2017partial} & \multicolumn{1}{c|}{$0.563$}  \\
\hline
Rampini \textit{et al.} \cite{rampini2019correspondence} & \multicolumn{1}{c|}{$0.676$}  \\
\hline
\textbf{Proposed dual spectra method} & \multicolumn{1}{c|}{$\boldsymbol{0.757}$}  \\ 
\hline
\end{tabular}
\caption{Results of the proposed method, the Bag-of-words baseline \cite{toldo2009bag} and recent axiomatic approaches \cite{rampini2019correspondence, rodola2017partial} tested on SHREC’16 (CUTS) \cite{cosmo2016shrec}. 
The proposed method outperforms all the competing axiomatic frameworks and achieves state-of-the-art results.}
\label{table: comparison}
\end{center}
\end{table}

\paragraph{Datasets. } We evaluate the proposed method on two databases.
The first is the \textbf{SHREC'16 Partial Matching Benchmark (CUTS)}, a standard dataset used to assess partial non-rigid shape matching frameworks.
The benchmark contains 120 partial shapes from 8 classes (dog, horse, wolf, cat, centaur, and 3 human subjects). 
The partial shapes are obtained by cutting full shapes that have undergone various non-rigid transformations with random planes.
The second is \textbf{FAUST-CUTS}, a dataset based on FAUST \cite{bogo2014faust}, that we introduce to assess the performance of partial shape matching frameworks in a cross-database configuration.
Our intention is to permit a fair comparison between different methods, by introducing new shapes, previously unseen by any learning method, with possibly different discretizations. 
FAUST-CUTS contains $25$ partial shapes from $10$ human subjects.
We refer the reader to the Supplementary for further details on the dataset and its construction.

\begin{figure}[htbp]
    \centering
    \includegraphics[width=0.5\textwidth]{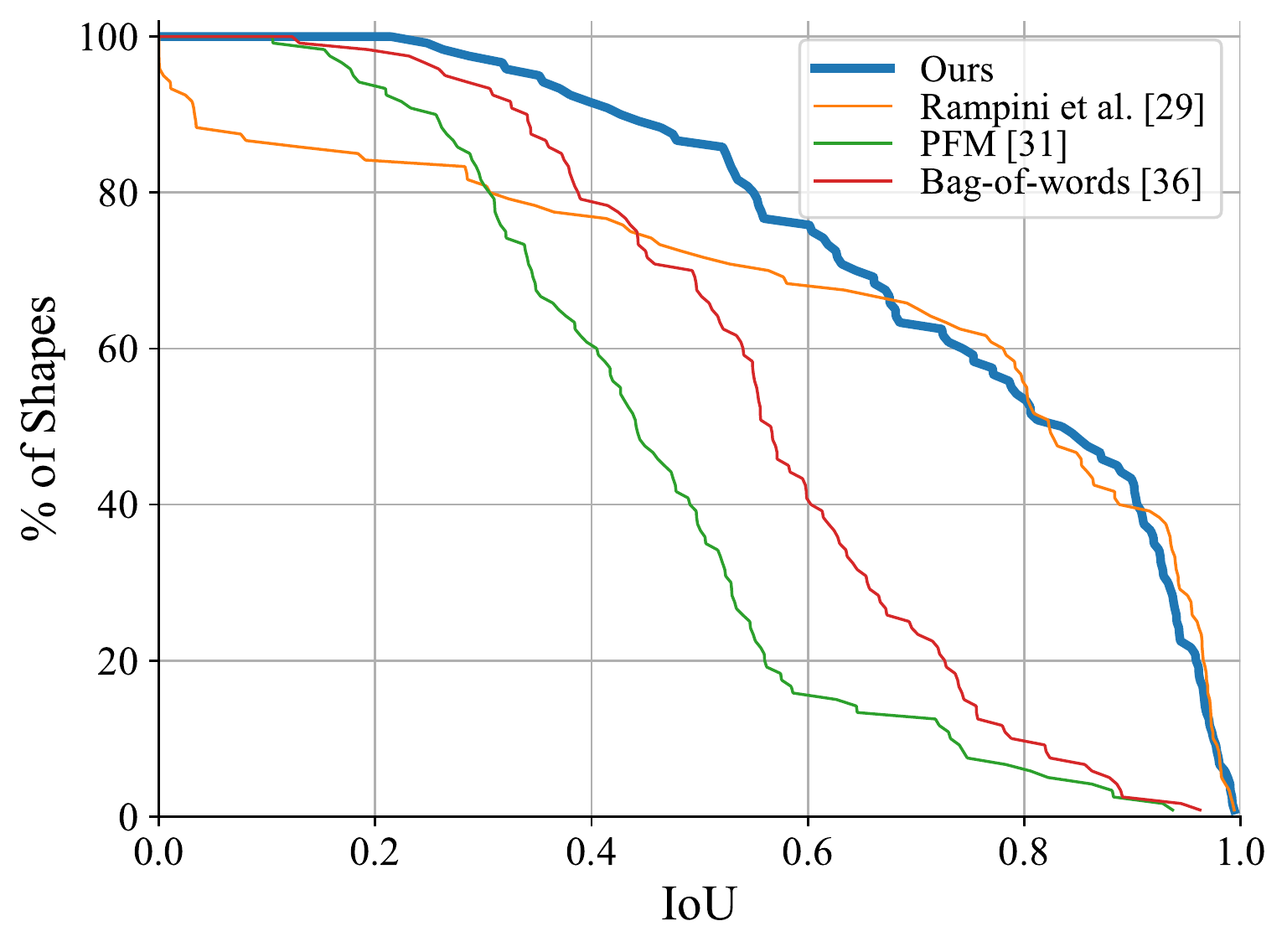}
    \caption{Comparison of the proposed method with recent frameworks \cite{rodola2017partial, rampini2019correspondence} and a bag-of-words \cite{toldo2009bag} of SHOT descriptors \cite{salti2014shot} as a baseline. The graph shows the cumulative score of each method over SHREC'16 (CUTS).}
    \label{fig: comparison}
\end{figure}

\begin{figure}[htbp]
    \centering
    \includegraphics[width=0.5\textwidth]{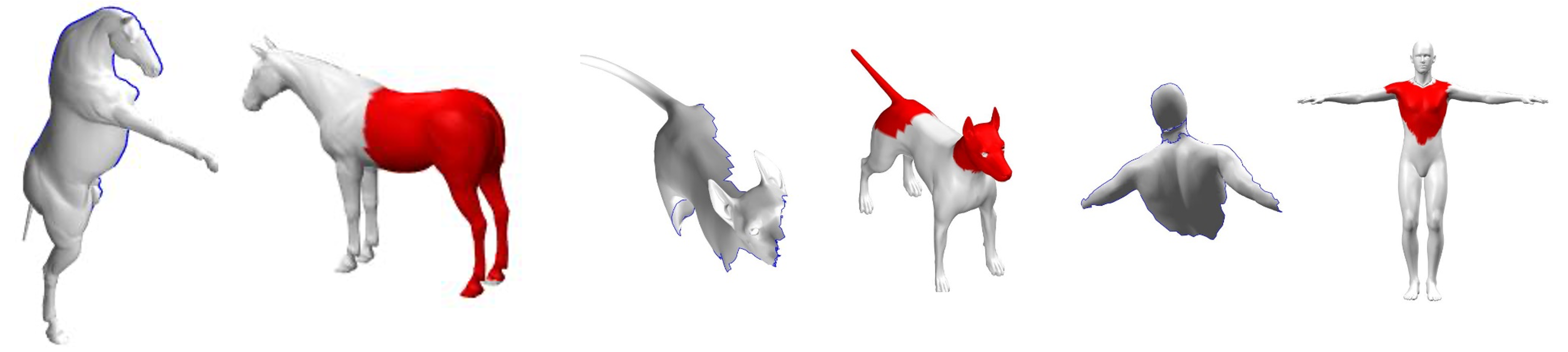}
    \caption{Analysis of failure cases. 
    \textbf{Left}: For quadrupeds, the distinction between cuts that contain both back or front legs, versus cuts that contain a back and a front leg, is a challenging task.
    \textbf{Middle}: Partial success on a challenging cut.
    \textbf{Right}: Human back and torso have close LBO and SI-LBO spectra. As a result, the proposed algorithm converges to a wrong local minimum.}
    \label{fig: Failure cases}
\end{figure}

\paragraph{Results.} We compare the proposed method with recent axiomatic frameworks for partial similarity on SHREC'16 (CUTS) \cite{cosmo2016shrec}.
We consider the spectrum alignment procedure proposed by Rampini \textit{et al.} \cite{rampini2019correspondence}, Partial Functional Maps (PFM) \cite{rodola2017partial} and a Bag-of-words approach \cite{toldo2009bag} with SHOT descriptors \cite{salti2014shot}.
We use the original code released by the authors and apply the best reported hyper-parameters.
Table \ref{table: comparison} and Figure \ref{fig: comparison} display a quantitative comparison, showing that the proposed method achieves significantly better results than other axiomatic methods.
Figure \ref{fig: qualitative results} presents successful examples of the proposed method on shapes from SHREC'16 \cite{cosmo2016shrec}.
Figure \ref{fig: Failure cases} analyzes failure cases that mostly arise due to similarity of distinct parts of human and animal bodies such as the torso and back (human) or front and back legs (quadrupeds).

\begin{figure}[htbp]
    \centering
    \includegraphics[width=0.5\textwidth]{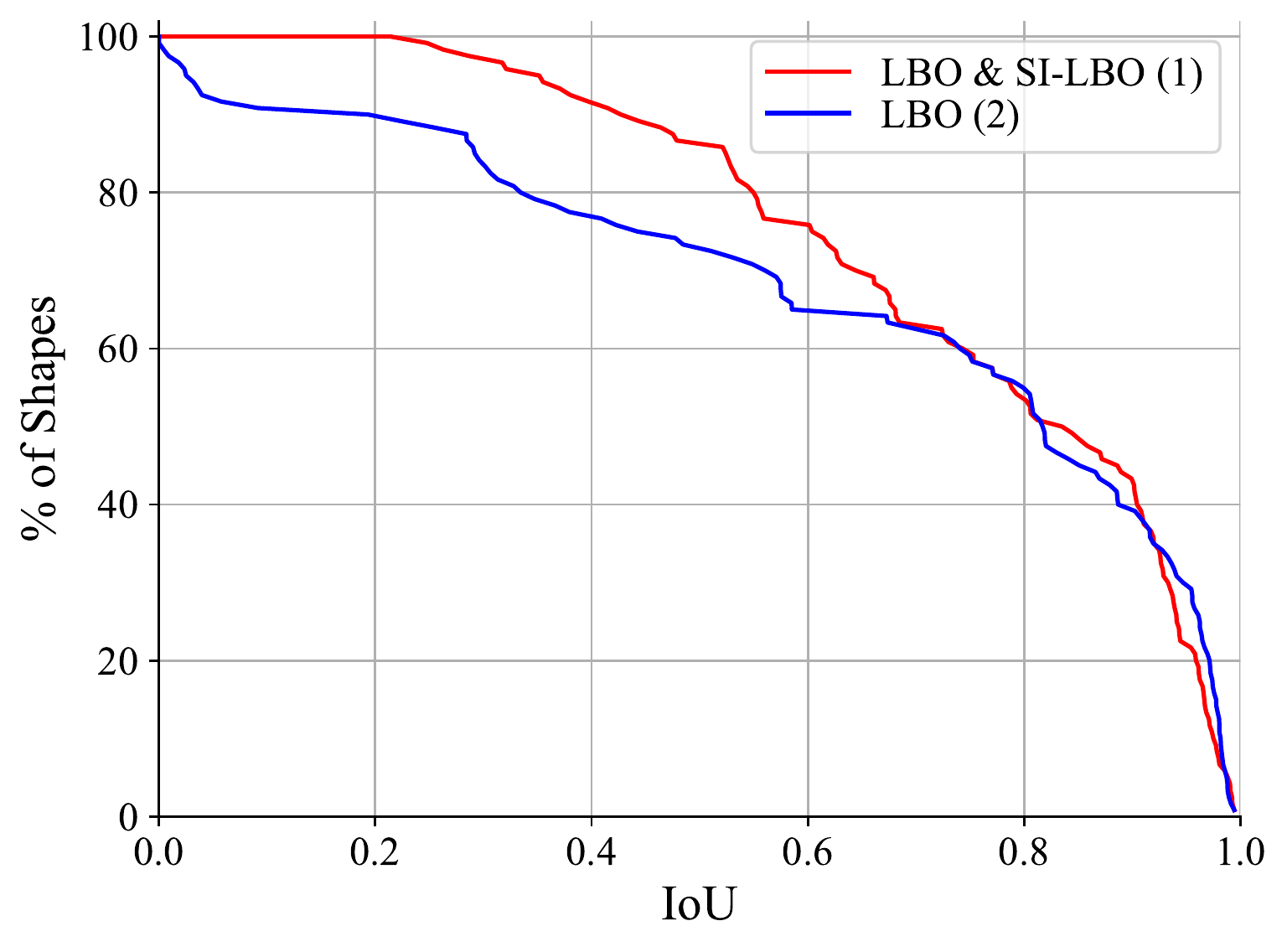}
    \caption{Ablation study. Comparison of the proposed multi-metric framework (1) with a truncated version of the proposed method which only relies on the LBO and its spectrum (2). The graph presents the cumulative scores of (1) and (2) over SHREC’16 (CUTS). (1) reaches a mean IoU of 0.76 and (2) a mean IoU of 0.71.}
    \label{fig: 40vs20+20}
\end{figure}

\paragraph{Ablation study.} 
We compare the performance of the proposed framework when considering (1) 20 eigenvalues of the LBO and 20 eigenvalues of the SI-LBO (2) 40 eigenvalues of the LBO (without including the SI-LBO).
A comparison of results obtained by (1) and (2) is displayed in Figure \ref{fig: 40vs20+20}. 
Note that for a fair comparison between these methods, an equal total number of eigenvalues is required. 
The region search is thus performed via alignment problems that have the same number of constraints.
The significantly better performance of (1) highlights the benefits of the proposed multi-metric approach and of the introduction of the scale-invariant metric.

\begin{figure}[htbp]
    \centering
    \includegraphics[width=0.5\textwidth]{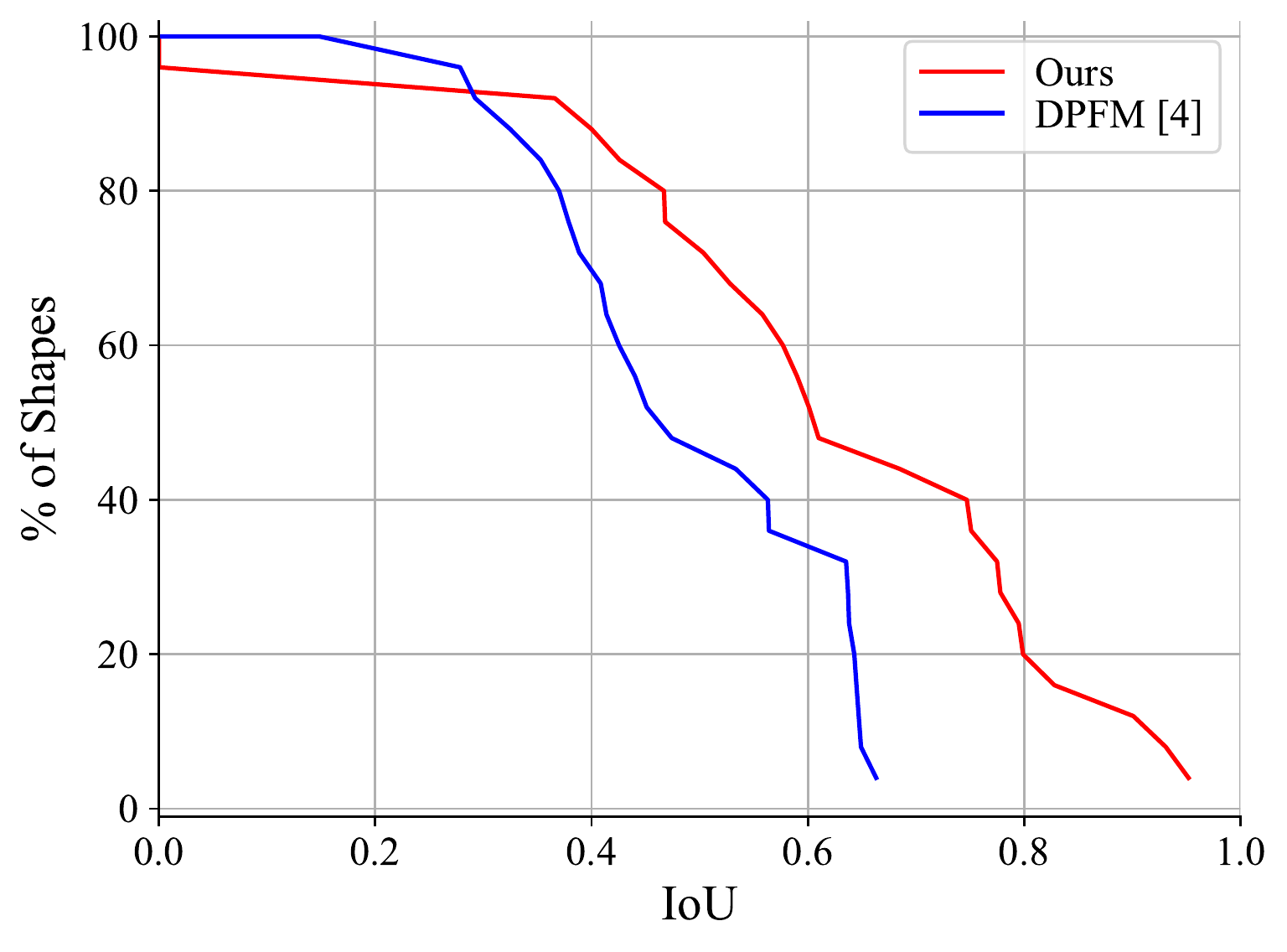}
    \caption{Comparison of the proposed method with Deep Partial Functional Map (DPFM) \cite{attaiki2021dpfm} on the newly introduced FAUST-CUTS dataset.
    DPFM is the state-of-the-art learning method. It is trained on SHREC'16. The proposed axiomatic approach significantly outperforms DPFM on the FAUST-CUTS dataset. It demonstrates that the proposed method enjoys better generalization properties.}
    \label{fig: generalization}
\end{figure}

\paragraph{Cross-dataset generalization \& comparison with learning methods.} We use FAUST-CUTS to compare the generalization ability of the proposed data-agnostic method with Deep Partial Functional Map (DPFM) \cite{attaiki2021dpfm}, the state-of-the-art learning method, trained on SHREC'16.
Figure \ref{fig: generalization} shows that the proposed method significantly outperforms DPFM and displays a clear advantage over actual learning approaches when processing new databases with unfamiliar shapes.

Additional experiments and implementation details of the proposed method can be found in the Supplementary.
\section{Future research directions}
We introduced a new approach for partial shape similarity.
The proposed method exploits the complementary perspective provided by the scale-invariant metric to significantly improve the performance reached by alternative methods.
One future research direction is the extension of the proposed method to more challenging datasets such as the SHREC’16 Partial Matching Benchmark (HOLES) \cite{cosmo2016shrec}. 
This could notably be done by considering new metric spaces as well as other differentiable shape representations that also adopt a multi-metric approach, such as the self-functional maps \cite{halimi2018self}.
Finally, the proposed method is fully differentiable and can potentially be used as an \textit{unsupervised loss} to improve deep learning setups for partial shape matching.
\paragraph{Acknowledgement.} We would like to thank Alon Zvirin for his assistance.


\bibliography{bib.bib}
\bibliographystyle{plain}
\end{document}


\title{Partial Shape Similarity via Alignment of Multi-Metric Hamiltonian Spectra - Supplement}
\def\httilde{\mbox{\tt\raisebox{-.5ex}{\symbol{126}}}}
\author{David Bensa\"id \\ Technion - Israel Institute of Technology \\ \tt\small dben-said@campus.technion.ac.il  \and Amit Bracha \\ Technion - Israel Institute of Technology \\ \tt\small amit.bracha@cs.technion.ac.il   \and Ron Kimmel \\ Technion - Israel Institute of Technology \\
\tt\small ron@cs.technion.ac.il}
\maketitle
We present additional results and analyses. 
Specifically,  
 we present simple examples illustrating the scale-invariant metric as well as a formulation and a proof of Property 4 in the continuous case. 
Then, a deeper analysis of the modes of the scale-invariant Laplace-Beltrami operator is proposed.
Next, key implementation considerations are reviewed followed by 
 further experiments and a description of FAUST-CUTS, the new dataset introduced to evaluate partial shape matching in a cross-dataset configuration.
Finally, 
 the self-functional map framework is presented as a potential multi-metric alternative to our shape descriptor.

\section{Scale-invariant metric}
\label{sec: si}
\noindent \textbf{Scale-invariant metric for curves.} We first examine the case of a parametric curve $C: \Omega \subseteq \mathbb{R} \rightarrow \mathbb{R}^2$ scaled by a factor $\alpha > 0$.
The scale-invariant metric is based on the curvature $\kappa$ that measures the change in direction of a curve over an infinitesimal distance.
Intuitively, the curvature is related to the \textit{bending} of the curve and is defined via the radius $\rho$ of the {\it osculating} circle that best approximates the curve locally.  
 It holds,
\begin{eqnarray} 
\kappa_{C} = \frac{1}{\rho_{C}} \,.
\end{eqnarray}
From simple geometric considerations, the radius $\rho_{(\alpha C)}$ in the scaled curve $(\alpha C)$ is equal to the radius $\rho_{C}$ in $C$ multiplied by ${\alpha}$.
Therefore,
\begin{eqnarray}
    \kappa_{(\alpha C)} &=& \frac{1}{\rho_{(\alpha C)}} \,=\, \frac{1}{\alpha} \kappa_{C} \,.
\label{eq: scaled-k}
\end{eqnarray}
\begin{theorem}
Let $C: \Omega \rightarrow \mathbb{R}^2$ be a smooth parametric curve. 
The length element of $C$ defined by $d\tau=|\kappa(s)| ds$, where $s$ is the Euclidean arc-length, is scale invariant.
\end{theorem}
\noindent \textbf{Proof.}
First remark that the parametrization $ds$ respects,
\begin{equation}
    ds_{(\alpha C)} = |(\alpha C)'| dp = |\alpha||C'| dp = \alpha ds_{C} \,.
\label{eq: scaled-ds}
\end{equation}
where $p$ is a general parametrization of $C$. 
Using Eq. (\ref{eq: scaled-ds}) and Eq. (\ref{eq: scaled-k}), 
\begin{eqnarray}
    d\tau_{(\alpha C)} = |\kappa_{(\alpha C)}| ds_{(\alpha C)} = \frac{|\kappa_{ C}|}{\alpha} \alpha ds_{C} = d\tau_{C} \,.
\end{eqnarray}
\paragraph{Extension to surfaces.}
We extend the scale-invariant metric introduced on curves to a smooth parametric surface $S$. 
The \textit{normal curvature} of a curve $C$ on $S$ is defined as $\langle \kappa \hat{n}, \hat{N}\rangle \,$, where $\hat{n}$ stands for the normal of $C$ and $\hat{N}$ for the normal of $S$.
We can parameterize all the curves passing through a point of $S$ with an orientation $\theta$ measured with respect to an arbitrary direction. 
The normal curvature $\kappa_n(\theta)$ admits a maximum $\kappa_1$ and a minimum $\kappa_2$. 
Their product defines the intrinsic \textit{Gaussian curvature}, $K \,= \,{\kappa_1 \kappa_2} \,$.
The scale-invariant metric for surfaces is obtained by modulating the Euclidean arc length by the Gaussian curvature.
\subsection{Examples}
\begin{figure}[htbp]
    \centering
    \includegraphics[width=0.2\textwidth]{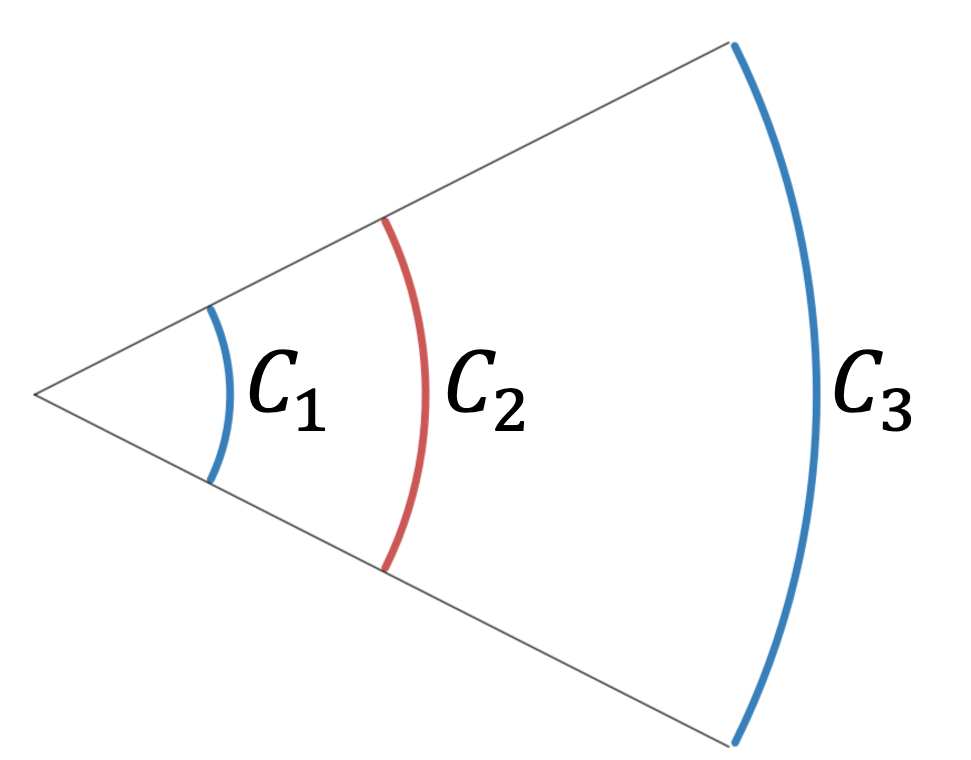}
    \caption{$C_1$ is an arc of the unit circle. 
    $C_2$ and $C_3$ are scaled versions of $C_1$ with factors of $2$ and $4$, respectively.
    $C_1$, $C_2$ and $C_3$ are \textbf{isometric} with respect to the scale-invariant metric and \textbf{non-isometric} with respect to the \textit{regular} one.
    }
    \label{fig: SI curve}
\end{figure}

\begin{figure}[htbp]
    \centering
    \includegraphics[width=0.5\textwidth]{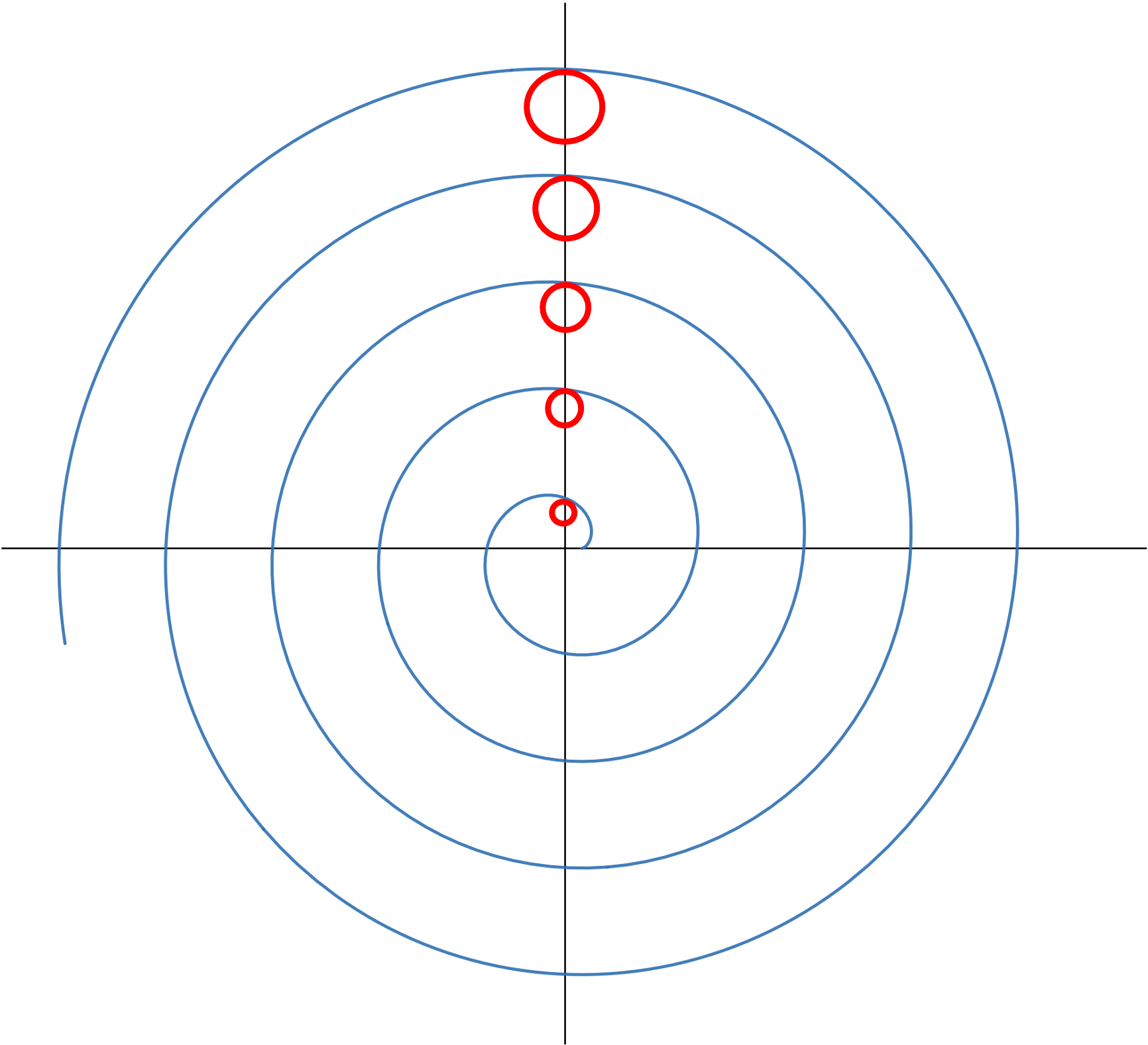}
    \caption{The well-known \textit{involute of a circle}. The radii of the osculating circles (in red) grow with the radius of the spiral. The scale-invariant arc-length is therefore constant over any full revolution ($2\pi$).}
    \label{fig: Involute of a circle}
\end{figure}

Here, we present simple examples showcasing the scale-invariant metric. 
Figure \ref{fig: SI curve} displays $3$ curves that are \textbf{isometric} with respect to the scale-invariant metric and \textbf{not isometric} with respect to the regular one.
\newline 
Figure \ref{fig: Involute of a circle} shows that the scale-invariant metric naturally scales different portions of the seminal \textit{involute of a circle}.
\paragraph{Involute of a circle.} The common planar formulation of the involute of the circle and its first order derivatives is:
\begin{eqnarray*}
    \begin{cases}
      x(t) = \cos(t) + t \, \sin(t)\\
      y(t) = \sin(t) - t \, \cos(t)
\end{cases}
\begin{cases}
    x_t = t \, \cos(t) \\
    y_t = t \, \sin(t) \,.
\end{cases}   
\end{eqnarray*}
When using the Euclidean arc length parametrization
$ds = t \, dt $, the second order derivatives are,
 \begin{eqnarray*}
 \begin{cases}
    x_{ss} = -\frac{\sin(t)}{t} \cr
    y_{ss} = \frac{\cos(t)}{t}\,.
\end{cases}  
\end{eqnarray*}
We finally obtain the following radius and curvature,
 \begin{eqnarray*}
 \begin{cases}
    \kappa(t) = \sqrt{\frac{\cos^{2}(t) + \sin^{2}(t)}{t^2}}  = \frac{1}{t}, \,\, (t>0)\\
    r(t) = t \, .
\end{cases}  
\end{eqnarray*}

\begin{figure*}[htbp]
    \centering
    \includegraphics[width=0.8\textwidth]{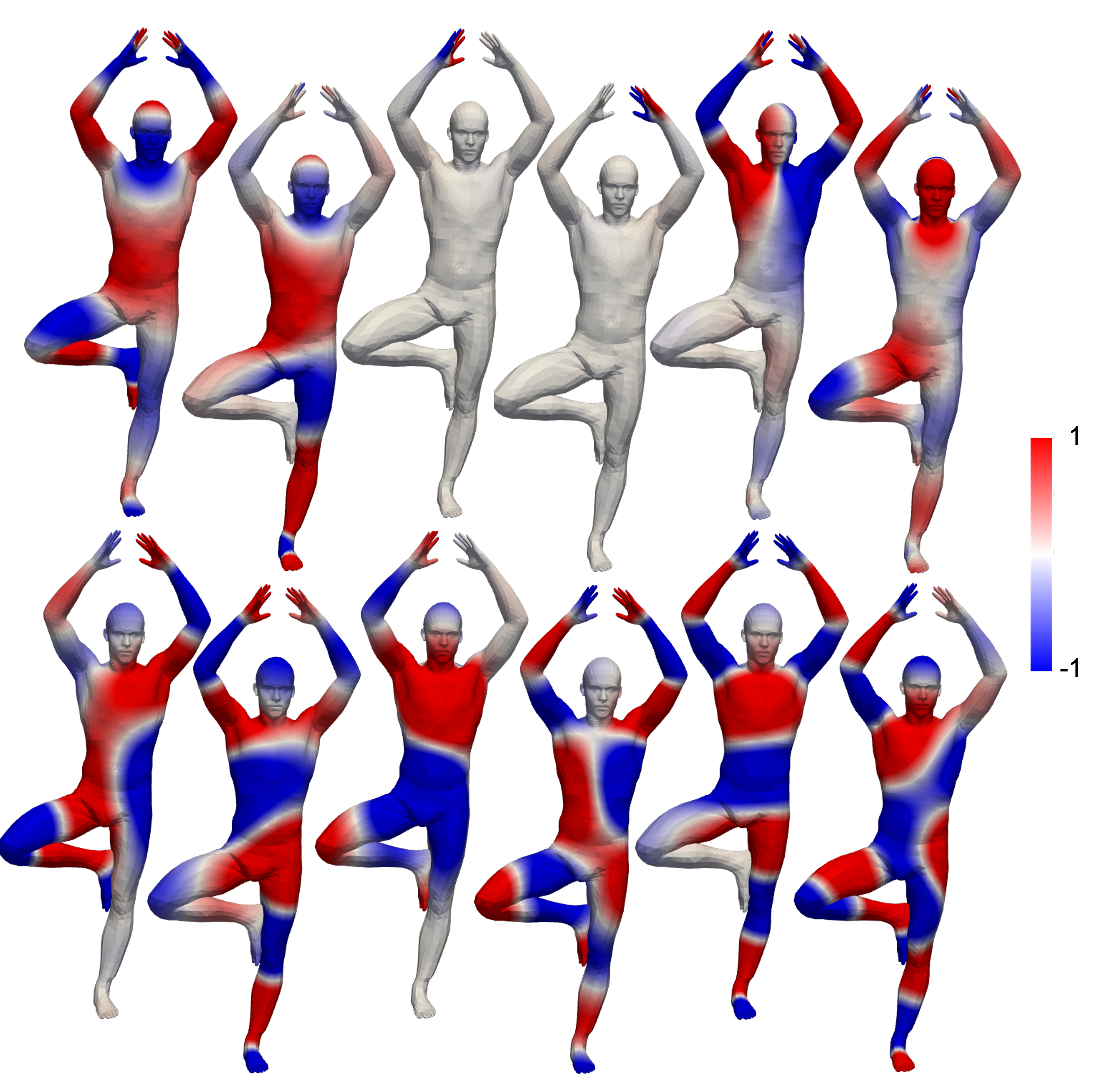}
    \caption{The $15^{th}$ to $20^{th}$ eigenvalues of the LBO (bottom) and the SI-LBO (top) of the $97^{th}$ shape from the Faust Dataset. 
    The eigenvalues of the $17^{th}$ \& $18^{th}$ modes are fully \textit{compressed} in the hands. Other modes of the SI-LBO are also concentrated in curved regions such as the fingers of the foot and the hands, and the head.}
    \label{fig: si-prior}
\end{figure*}

\subsection{Continuous formulation \& proof of the perturbation analysis proposed in Section 4.2}
\begin{theorem}
Consider a \textit{small} perturbation $\delta_p: \Omega \subseteq \mathbb{R}^2 \rightarrow \mathbb{R}^3$ of the curvature at $p \in \tilde{\mathcal{M}}$ such that,
\begin{eqnarray*} 
    \int_{\omega \subseteq \Omega} \delta_p(\omega) da &=& \begin{cases}
   \epsilon > 0,  & p \in \Omega \\
    0,  & {\text{otherwise}} \,.
\end{cases} 
\end{eqnarray*}
Denote by $\delta K \triangleq \frac{\delta_p}{\int_{\Omega}|\text{K}|da}$ the relative perturbation of the curvature and by $\delta \tilde{\lambda}  \triangleq \frac{\tilde{\lambda}_i - \tilde{\mu}_i}{\tilde{\lambda}_i}$ the relative perturbation of $\tilde{\lambda}_i$, where $\tilde{\mu}_i$ designates the $i^{th}$ eigenvalue of the perturbed manifold. $ \delta \tilde{\lambda}$ respects, 
\begin{eqnarray*}
    \delta \tilde{\lambda} \sim \delta K \,.
\end{eqnarray*}
\end{theorem}
\paragraph{Proof.}
From the generalization of the Weyl law proposed in Section 4,
\begin{eqnarray*}
\tilde{\lambda}_i \sim \frac{2\pi i}{\int_{\Omega}|\text{K}(\omega)| da}  \, ,\quad \tilde{\mu}_i \sim \frac{2\pi i}{\int_{\Omega}|\text{K}(\omega)| da + \epsilon}  \,.
\end{eqnarray*}
Therefore,
\begin{align*}
     \delta \tilde{\lambda} &\triangleq \frac{\tilde{\lambda}_i - \tilde{\mu}_i}{\tilde{\lambda}_i}  \cr
     &\sim \frac{1}{\tilde{\lambda}_i} \frac{2 \pi i \epsilon}{(\int_{\Omega}|\text{K}(\omega)| da)(\int_{\Omega}|\text{K}(\omega)| da + \epsilon)} \cr
    &\sim \frac{1}{\tilde{\lambda}_i} \frac{2 \pi i \epsilon}{(\int_{\Omega}|\text{K}(\omega)| da)^{2}} = \delta \textbf{K}  \,.
\end{align*}

\subsection{SI-LBO eigenfunctions and high frequencies}
Here, we extend the discussion of Section 4.2 on the eigenfunctions of the SI-LBO. 
Figure \ref{fig: si-prior} showcases that the basis induced by the spectral decomposition of the SI-LBO is empirically more concentrated, or \textit{compressed}, in curved regions. 
Local features, such as the fingers in the human case, are generally captured by modes corresponding to large eigenvalues ("\textit{high frequencies}") of the LBO.
By concentrating the main variations of its eigenfunctions in curved regions, the SI-LBO allows to capture key features with modes related to \textit{low frequencies}.  
This property is especially important in practical settings which rely on a \textbf{truncated} spectrum and where larger imprecisions are observed when computing large eigenvalues and eigenvectors.


\section{Implementation details}
\label{sec: implementation}
\paragraph{Initialization.} 
Following \cite{rampini2019correspondence}, multiple initialization of the potential $\text{V}$ are considered. 
We sample $20$ points on the full shape $\mathcal{M}$ with a farthest sampling strategy. 
Each sample defines the center of $2$ Gaussians with variances respectively equal to ${\sqrt{\text{Area}(\mathcal{M})}}$ and to $\sqrt{2 \, \text{Area}(\mathcal{M})}$.
The baseline solution consisting of a Bag-of-words \cite{toldo2009bag} of SHOT descriptors \cite{salti2014shot} is also considered as an initialization.
We adopt this initialization policy to fairly compare our method with the closely related work of Rampini \textit{et al.} \cite{rampini2019correspondence}.
The cost function is separately minimized for each initialization in parallel. 
The final solution is selected by comparing the projections of SHOT descriptors \cite{salti2014shot} on the first eigenfunctions of the \textit{regular} and of the scale-invariant Hamiltonian of $\mathcal{M}$, with the projection of SHOT descriptors on the first eigenfunctions of the LBO and of the SI-LBO of the partial shape $\mathcal{N}$.

\begin{figure}[htbp]
    \centering
    \includegraphics[width=0.5\textwidth]{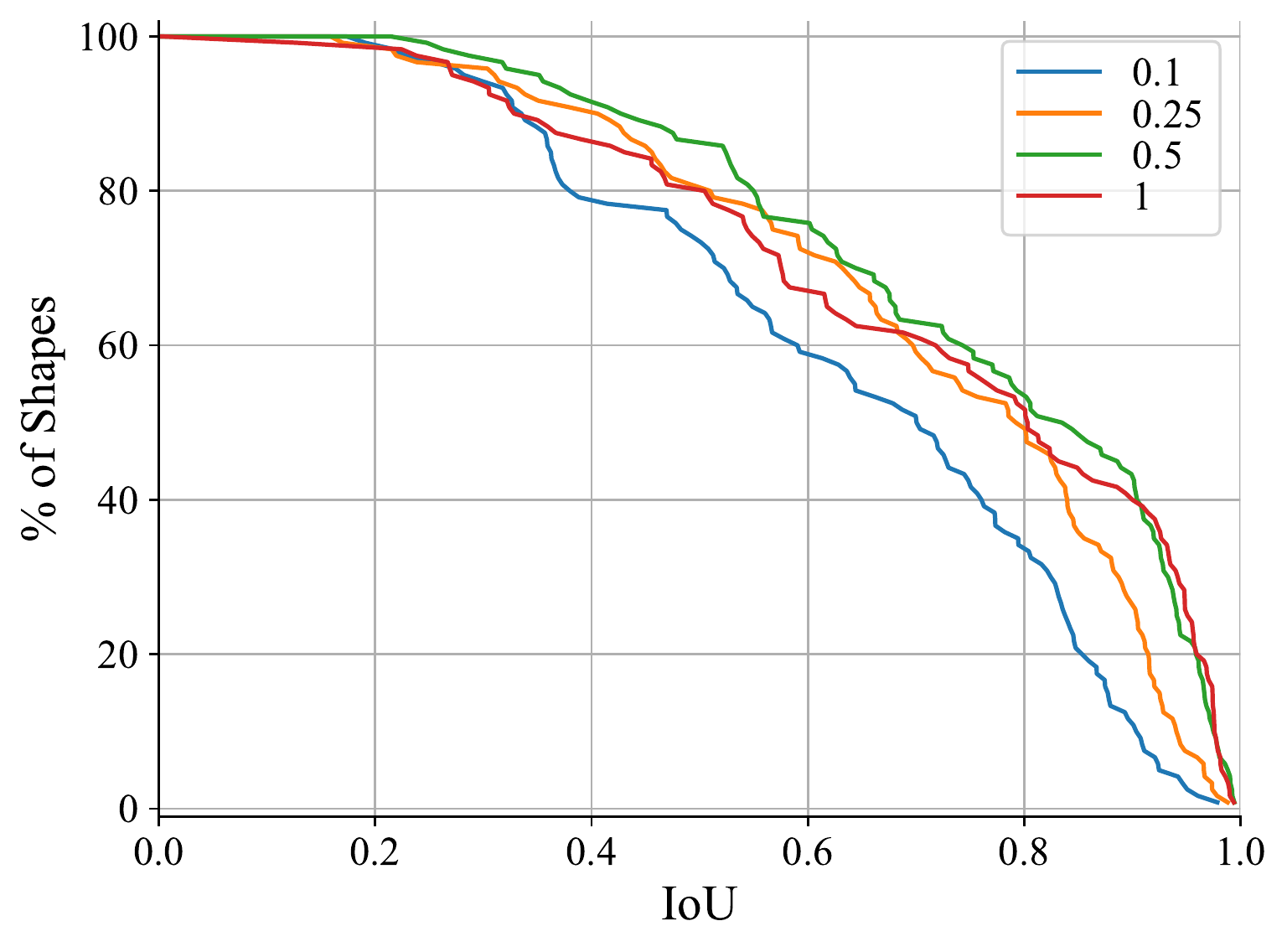}
    \caption{Cumulative IoU of the proposed method over shapes from SHREC'16 (CUTS) with different vertex densities. We used a mesh decimation procedure and each shape was resampled with 50\%, 25\%, and 10\% of the original number of vertices. Even when most of the shape vertices are decimated, only a moderate drop in performance is observed. This experiment demonstrates the robustness of our method to changes in vertex density and its ability to process shapes over a wide range of resolutions.}
    \label{fig: density}
\end{figure}

\paragraph{Scale-invariant metric.} 
Estimating Gaussian curvature on a triangular mesh is an active area of research \cite{surazhsky2003comparison}. 
The Gaussian curvature K is smoothed to overcome the discrepancies caused by the discretization of the shape: For each vertex we consider the mean value of the Gaussian curvature over its first ring neighbors.
Moreover, following \cite{aflalo2013scale, bracha2020shape, pazi2020unsupervised}, we use the metric,
\begin{equation*}
    (|K| + \epsilon)^{\alpha} g,
\end{equation*}
with $\alpha \in [0, 1]$, which is interpolated between the regular ($\alpha=0$) and the fully scale-invariant metric ($\alpha=1$).
Interestingly, the introduction of the parameter $\alpha$ is meaningful in light of the interpretation of the scale-invariant metric proposed in Section 4.2. $\alpha$ regulates the influence of the \textit{shape prior} and quantifies the importance given to features found in curved regions.
For all the experiments we considered $\alpha = 0.33$ and $\epsilon=10^{-8}$.
\newline
\textbf{An implementation of our framework will be publicly released after publication.}
\section{Experiments}
\label{sec: experiments}
\subsection{Robustness to different discretizations}
Figure \ref{fig: density} shows the robustness of our method to modification of the resolution of the processed meshes. In contrast with frameworks relying on local descriptors such as PFM \cite{rodola2017partial}, our method is based on a global shape descriptor and is robust to different tessellations. 
This experiment also displays the effectiveness of our approach on meshes with low resolution. 

\subsection{Length of the truncated spectrum}
\begin{figure}[htbp]
    \centering
    \includegraphics[width=0.5\textwidth]{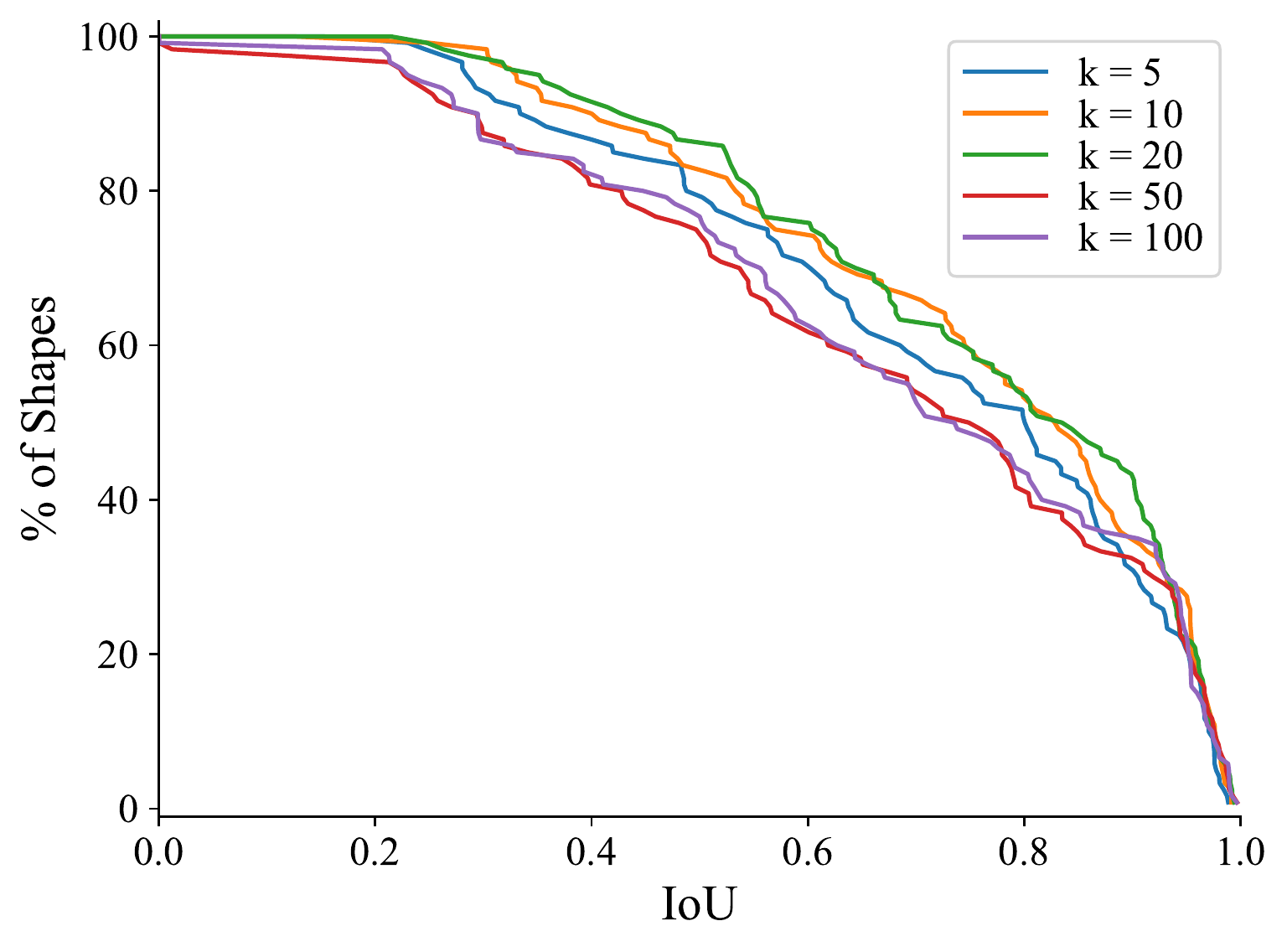}
    \caption{Cumulative IoU of the proposed method over SHREC'16 (CUTS) with different truncated spectra of different lengths.}
    \label{fig: truncated}
\end{figure}
Figure \ref{fig: truncated} displays an evaluation of the performance of our method when considering different number of eigenvalues. 
As expected, the transition from small truncated spectra, including $5$ or $10$ eigenvalues, to truncated spectra with $20\!-\!30$ eigenvalues improves the results obtained.
Interestingly, the  performance decreases for truncated spectra including more than $k=40$ eigenvalues. 
This can be explained by the inaccurate estimation of large eigenvalues and by the heavier computational cost of the alignment of long spectra.

\subsection{FAUST-CUTS}
\begin{figure}[htbp]
    \centering
    \includegraphics[width=0.5\textwidth]{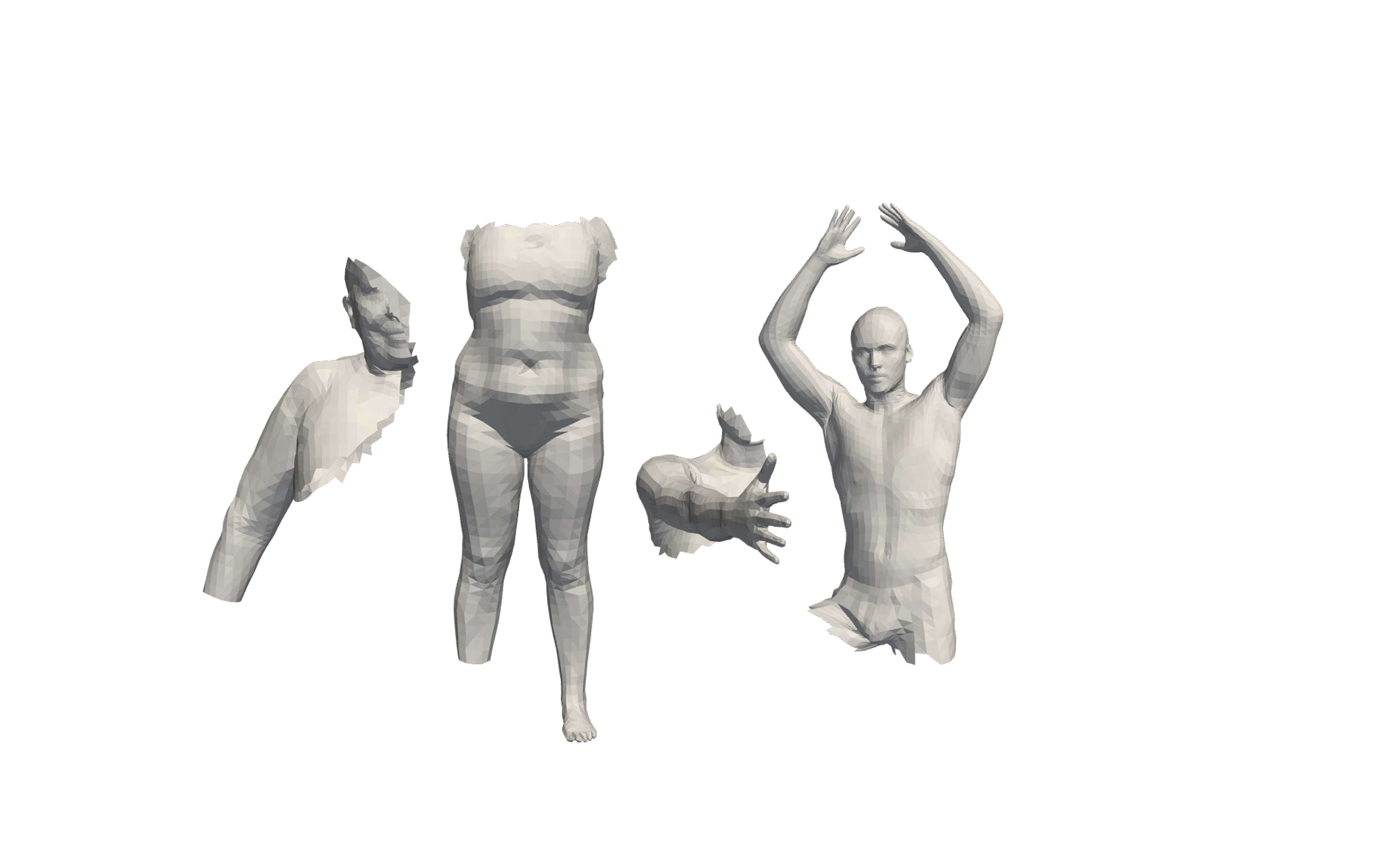}
    \caption{Partial shapes from FAUST-CUTS.}
    \label{fig: FAUST}
\end{figure}

\textbf{FAUST-CUTS} is a dataset for partial shape matching based on FAUST \cite{bogo2014faust}. 
It was introduced to assess the performance of partial shape matching frameworks in a cross-database configuration. 
FAUST-CUTS contains $25$ partial shapes from $10$ human subjects.
The partial shapes are obtained by cutting full shapes that have undergone various non-rigid transformations.
\paragraph{Dataset construction.} Shapes from different subjects and in different poses are randomly selected. 
Then, we sample $10$ points per shape using a farthest sampling strategy.
One of those points and a radius (between $20\%$ and $70\%$ of the shape radius) are randomly selected. 
They define a partial shape that should be located on a full shape of the corresponding subject in a neutral pose.
Figure \ref{fig: FAUST} displays examples of partial shapes from FAUST-CUTS.
Meshes from FAUST and TOSCA are different in terms of resolution and tessellation.
FAUST-CUT is thus an interesting benchmark to evaluate the generalization ability and the robustness of learning methods trained on TOSCA.

\section{Self-functional map, an alternative multi-metric shape representation}
\label{sec: Self-functional map}
In 2018, Halimi \textit{et al.} \cite{halimi2018self} introduced a compact shape signature named \textit{Self-functional map}.
The Self-functional map framework encodes the interactions between the eigenfunctions of Laplace-Beltrami operators defined on two different Riemannian manifolds of the same surface. 
Namely, the Self-functional map $\text{D} \in R^{k \times k}$ is,
\begin{eqnarray*}
D_{ij} = \langle \psi_i , \phi_j \rangle_{w} \,,
\end{eqnarray*}
where ${\{\psi\}_{i=1}^{k}}$ and ${\{\phi\}_{i=1}^{k}}$ are two truncated bases composed of the eigenfunctions of the LBOs defined on \textit{different} metric spaces. $w$ is the metric tensor that induced $\{\psi\}_{i=1}^{k}$ or $\{\phi\}_{i=1}^{k}$.

The Self-functional map is closely related to our shape descriptor. 
It relies on the \textit{eigenfunctions} instead of the \textit{eigenvalues} of differential operators to create a compact shape signature.
\begin{theorem}{The eigenfunctions $\{ \phi_i \}$ of the discretized Hamiltonian operator $H = \Delta_{\mathcal{M}} + v$ are differentiable according to ${v}$. 
Namely,
\begin{eqnarray*}
\frac{\partial \phi_i}{\partial v} &=&   - \Phi (\textup{diag}(\lambda) - \lambda_i I)^{+}  \Phi^T \textbf{A} \phi_i 1^T,
\end{eqnarray*}
where ${\lambda_i}$ stands for the eigenvector corresponding to $\phi_i$ and $\Phi$ a column-matrix containing the truncated basis considered.}
\label{theorem: eigenfunctions derivation}
\end{theorem}
\paragraph{Proof.}
According to \cite{abou20091}, 
\begin{eqnarray*}
    \frac{\partial}{\partial v} \phi_i &=& - \Phi (\text{diag}(\lambda) - \lambda_i I)^{+}  \Phi^T \frac{\partial}{\partial v} (\textbf{W} + \textbf{A} \, \text{diag}(\textbf{V})) \phi_i  \cr &=& - \Phi (\text{diag}(\lambda) - \lambda_i I)^{+}  \Phi^T \textbf{A} [\phi_i, ...\phi_i]_{n \times n}  \cr 
    &=& - \Phi (\text{diag}(\lambda) - \lambda_i I)^{+}  \Phi^T \textbf{A} \phi_i 1^T \,
\label{eq: }
\end{eqnarray*}

$\text{with} \, \, \phi_i^T  A \phi_i = I$. 
The Monroe pseudo inverse $+$ can be efficiently computed in our case since ${(\text{diag}(\lambda) - \lambda_i I)}$ is a diagonal matrix.
Here, the introduction of a novel metric allows to elegantly use the information contained in the eigenfunctions in a global manner - without point-wise quantities.
The last observation together with Theorem \ref{theorem: eigenfunctions derivation} make possible the resolution of tasks such as shape similarity or shape retrieval from the Self-functional map embedding.
This direction should be explored in a future research. 
The Self-functional map framework illustrates the exciting opportunities opened by the introduction of multiple bases and the potential of shape analysis methods joining the complementary information offered by different metric spaces.

\bibliography{bib.bib}
\bibliographystyle{plain}